\pdfoutput=1

\documentclass[11pt]{article}

\usepackage{acl}

\usepackage{times}
\usepackage{latexsym}

\usepackage[T1]{fontenc}

\usepackage[utf8]{inputenc}

\usepackage{microtype}

\usepackage{inconsolata}

\usepackage{graphicx}
\usepackage{placeins}

\usepackage{pifont} 
\usepackage[dvipsnames]{xcolor}
\newcommand{\cmark}{\textcolor{OliveGreen}{\ding{51}}} 
\newcommand{\xmark}{\textcolor{BrickRed}{\ding{55}}} 

\usepackage[many]{tcolorbox} 
\definecolor{blue}{HTML}{5989cf}
\definecolor{sub}{HTML}{cde4ff}
\newtcolorbox{todobox}{
    colback = sub, 
    colframe = blue, 
    boxrule = 0pt, 
    leftrule = 6pt 
}

\usepackage[edges]{forest}

\usepackage{caption}
\usepackage{float}

\title{Bring Your Own Knowledge: A Survey of Methods\\ for LLM Knowledge Expansion}
\author{
Mingyang Wang$^{1,2,3}$\thanks{Equal contribution.} \hspace*{0.2cm}
Alisa Stoll$^{4}$\footnotemark[1] \hspace*{0.2cm}
Lukas Lange$^{1}$ \\
{\bf Heike Adel$^{5}$ \hspace*{0.2cm} Hinrich Sch\"{u}tze$^{2,3}$ \hspace*{0.2cm} Jannik Str\"{o}tgen$^{4}$} \\
$^1$Bosch Center for Artificial Intelligence (BCAI) \hspace*{0.2cm}
$^2$LMU Munich \\
$^3$Munich Center for Machine Learning (MCML) \\
$^4$Karlsruhe University of Applied Sciences \\
$^5$Hochschule der Medien, Stuttgart\\
\texttt{mingyang.wang2@de.bosch.com}
}
\begin{document}
\maketitle
\begin{abstract}
Adapting large language models (LLMs) to new and diverse knowledge is essential for their lasting effectiveness in real-world applications. This survey provides an overview of state-of-the-art methods for expanding the knowledge of LLMs, focusing on integrating various knowledge types, including factual information, domain expertise, language proficiency, and user preferences. We explore techniques, such as continual learning, model editing, and retrieval-based explicit adaptation, while discussing challenges like knowledge consistency and scalability. Designed as a guide for researchers and practitioners, this survey sheds light on opportunities for advancing LLMs as adaptable and robust knowledge systems.
\end{abstract}



\section{Introduction}

As large language models (LLMs) are increasingly deployed in real-world applications, their ability to adapt to evolving knowledge becomes crucial for maintaining relevance and accuracy. However, LLMs are typically trained once and thus only have knowledge up to a certain cutoff date, limiting their ability to stay updated with new information. This survey provides a comprehensive overview of methods that enable LLMs to incorporate various types of new knowledge, including factual, domain-specific, language, and user preference knowledge. We survey adaptation strategies, including continual learning, model editing, and retrieval-based approaches, and aim at providing guidelines for researchers and practitioners.

To remain effective, LLMs require updates across multiple dimensions. Factual knowledge consists of general truths and real-time information, while domain knowledge pertains to specialized fields, such as medicine or law. Language knowledge enhances multilingual capabilities, and preference knowledge aligns model behavior with user expectations and values. Ensuring that LLMs can integrate updates across these dimensions is essential for their sustained utility.

\begin{table}[t]
\scalebox{0.8}{
\begin{tabular}{l|c|c|c}
\hline
& Continual & Model & \\
& Learning  & Editing & Retrieval \\
& (§\ref{sec:continual-learning}) &
(§\ref{sec:model-editing}) &
(§\ref{sec:retrieval}) \\ \hline
\textbf{Knowledge Type}    &  &  & \\ 
Fact           & \cmark & \cmark & \cmark \\
Domain         & \cmark & \xmark & \cmark \\
Language       & \cmark & \xmark & \xmark \\
Preference     & \cmark & \cmark & \xmark                    \\ \hline

\textbf{Applicability}    &  &  & \\ 
Large-scale data   & \cmark & \xmark & \cmark \\ 
Precise control & \xmark & \cmark & \cmark \\ 
Computational cost    & \xmark & \cmark & \cmark \\ 
Black-box applicable    & \xmark & \xmark & \cmark \\ 

\hline
\end{tabular}}
\caption{We compare three key approaches for adapting LLMs --- continual learning, model editing, and retrieval --- based on their supported knowledge types and applicability across different criteria.}
\label{tab:method-summary}
\end{table}

Existing LLM adaptation methods differ in approach and application. Continual learning enables incremental updates to models’ parametric knowledge, mitigating catastrophic forgetting \citep{mccloskey1989catastrophic} while ensuring long-term performance. Model editing allows for precise modifications of learned knowledge, providing controlled updates without requiring full retraining. Unlike these \textit{implicit} knowledge expansion methods, which modify the model’s internal parameters, retrieval-based approaches \textit{explicitly} access external information dynamically during inference, reducing dependency on static parametric knowledge. The suitability of these methods for different knowledge types and their general applicability are summarized in Table~\ref{tab:method-summary}. By leveraging these strategies, LLMs can maintain accuracy, contextual awareness, and adaptability to new information. 


\tikzset{
    every node/.style={
        align=left,
    }
}
\begin{figure*}[ht]
\centering
\resizebox{\textwidth}{!}{
    \begin{forest}
    for tree={
        text width=3.5cm,
        grow=east,
        parent anchor=east,
        child anchor=west,
        anchor=center,
        rounded corners,
        font=\footnotesize,
        edge={thick},
        forked edges,
        draw,
        l sep=5mm
    }
    [Knowledge Expansion Methods,
        rotate=90,
        text width=4.2cm,
        draw=gray!50, 
        line width=0.5mm, 
        fill=gray!20,
        [
            Retrieval-based Methods (§\ref{sec:retrieval}), 
            text width=4.0cm,
            draw=yellow!50, 
            line width=0.5mm, 
            fill=yellow!20
            [
                Domain Adaptation (§\ref{sec:retrieval-domain}),
                text width=5.1cm,
                draw=yellow!50, 
                line width=0.5mm, 
                fill=yellow!20,
            ]
            [
                Updating Facts (§\ref{sec:retrieval-facts}),
                text width=5.1cm,
                draw=yellow!50, 
                line width=0.5mm, 
                fill=yellow!20,
            ]
        ]
        [
            Model Editing (§\ref{sec:model-editing}), 
            text width=4.0cm,
            draw=red!50, 
            line width=0.5mm, 
            fill=red!20 
            [
                Updating Preferences (§\ref{sec:model-editing-preference}), 
                text width=5.1cm,
                draw=red!50, 
                line width=0.5mm, 
                fill=red!20,
            ]
            [
                Updating Facts (§\ref{sec:model-editing-facts}), 
                text width=5.1cm,
                draw=red!50, 
                line width=0.5mm,
                fill=red!20,
            ] 
        ]
        [
            Continual Learning (§\ref{sec:continual-learning}), 
            text width=4.0cm,
            draw=orange!50, 
            line width=0.5mm, 
            fill=orange!20 
            [
                Continual Preference Alignment (§\ref{sec:continual-alignment}), 
                text width=5.1cm,
                draw=orange!50, 
                line width=0.5mm, 
                fill=orange!20,
            ]
            [
                Continual Pretraining, 
                text width=5.1cm,
                draw=orange!50, 
                line width=0.5mm, 
                fill=orange!20, 
                [
                    Language Expansion (§\ref{sec:continual-pretraining-language}),
                    text width=4.2cm,
                    draw=orange!50, 
                    line width=0.5mm, 
                    fill=orange!20,
                    yshift=-8.45mm, 
                    [
                        Programming Language,
                        text width=3.2cm,
                        draw=orange!50, 
                        line width=0.5mm, 
                        fill=orange!20,
                        yshift=-8.45mm, 
                    ]
                    [
                        Natural Language,
                        text width=3.2cm,
                        draw=orange!50, 
                        line width=0.5mm, 
                        fill=orange!20,
                        yshift=-8.45mm, 
                    ]
                ]
                [
                    Domain Adaptation (§\ref{sec:continual-pretraining-domain}),
                    text width=4.2cm,
                    draw=orange!50, 
                    line width=0.5mm, 
                    fill=orange!20,
                    yshift=-8.45mm, 
                ]
                [
                    Updating Facts (§\ref{sec:continual-pretraining-facts}),
                    text width=4.2cm,
                    draw=orange!50, 
                    line width=0.5mm, 
                    fill=orange!20, 
                    yshift=-8.45mm, 
                ],
            ] 
        ]
    ]
    \end{forest}
}
\caption{Taxonomy of current methods for expanding LLM knowledge. Due to space constraints, please refer to Appendix \ref{sec:appendix_taxonomy} for a comprehensive review of methods and their corresponding citations.}
\label{fig:taxonomy}
\end{figure*}
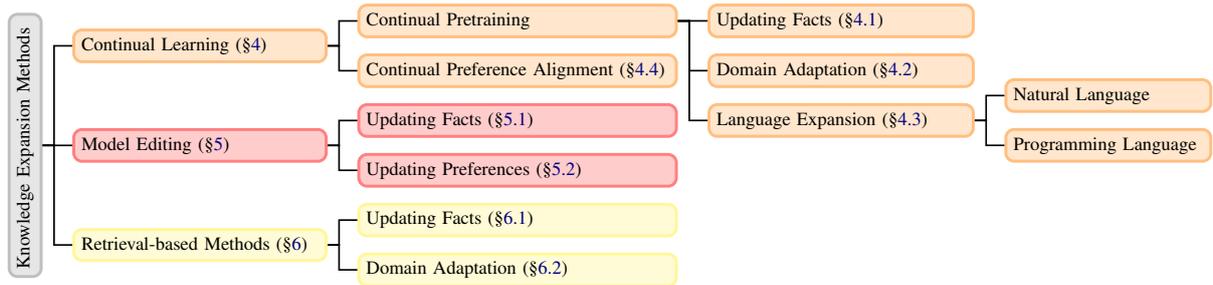

After placing our work into context 
(Section~\ref{sec:rel_surveys}) and defining  knowledge types covered in this paper (Section~\ref{sec:knowledge-types}), we provide an overview of different knowledge expansion methods as detailed in Figure~\ref{fig:taxonomy}.
This work thus surveys diverse research efforts and may serve as a guide for researchers and practitioners aiming to develop and apply adaptable and robust LLMs. We highlight research opportunities 
and provide insights into optimizing adaptation techniques for various real-world applications.





\section{Related Surveys}\label{sec:rel_surveys}
The main goal of our work is to provide researchers and practitioners a broad overview of various types of methods to adapt LLMs to diverse types of new knowledge. 
In this section, we explain how other more specialized surveys relate to our paper.

To the best of our knowledge, there is limited prior work that specifically focuses on continuous knowledge expansion for LLMs. 
Closest to our work, \citet{zhang-etal-2023-large} describe temporal factual knowledge updates, while we take a broader perspective by examining methods for adapting LLMs to unseen domain knowledge, expanding language coverage, and incorporating user preferences.
\citet{yao-etal-2023-editing} and \citet{zhang2024comprehensive} provide overviews of knowledge editing methodologies, categorizing approaches of knowledge editing. Similarly, \citet{ ke2022continual}, \citet{wu2024continual} and \citet{wang2024comprehensive} offer a comprehensive overview of continual learning. In contrast, our survey shifts the focus towards a task-oriented perspective on knowledge expansion, detailing how various types of knowledge --- including factual, domain-specific, language, and user preference knowledge --- can be seamlessly integrated to ensure LLMs remain relevant and effective. 



\section{Knowledge Types}\label{sec:knowledge-types}

Integrating diverse types of knowledge into LLMs is essential to enhance their versatility and effectiveness. Depending on the use case, the type of knowledge that an LLM shall be adapted to, might differ. In this paper, we distinguish four key types of knowledge, which cover a broad range of use cases of researchers and practitioners: \textbf{factual, domain, language, and preference knowledge.}

\textbf{(1)} We define \textbf{factual knowledge} as general truths or contextualized information about the world that can be expressed in factual statements. 
We adopt a broad, high-level definition, encompassing finer-grained categorizations, such as commonsense knowledge, cultural knowledge, temporal knowledge, and entity knowledge as subsets of factual knowledge, in contrast to prior works \citep{cao2024life, wu2024continual} using more granular classifications. This inclusive perspective enables a comprehensive exploration of knowledge expansion techniques for LLMs, providing flexibility beyond predefined categories and taxonomies.

\textbf{(2)} We define \textbf{domain knowledge} as specialized information relevant to specific fields, such as medicine, law, or engineering, enabling LLMs to perform well in targeted applications. Since LLMs typically excel in general-domain tasks but struggle with specialized content, incorporating domain knowledge is crucial for bridging this gap and improving performance in specific fields.

\textbf{(3)} We define \textbf{language knowledge} as the ability of an LLM to understand, generate, and reason in specific natural or programming languages.\footnote{We distinguish language knowledge from linguistic knowledge as defined by \citet{hernandez2024linearity}. Language knowledge refers to the multilingual capabilities of an LLM, whereas linguistic knowledge falls under factual knowledge, encompassing statements about syntax and grammar.} Its integration focuses on adapting models to new languages and enhancing performance in underrepresented ones for broader applicability.

\textbf{(4)} Finally, we define \textbf{preference knowledge} as the capability of LLMs to tailor their behavior to align with user-specific needs, preferences, or values. Preference knowledge integration involves adapting LLM behavior to meet diverse and dynamic user expectations.

In the next sections, we survey knowledge expansion methods and explain for which of these four knowledge types they are suitable.

\section{Continual Learning}
\label{sec:continual-learning}
Continual learning (CL) is a machine learning paradigm that mimics the human ability to continuously learn and accumulate knowledge without forgetting previously acquired information \citep{chen2018continual}. In the context of knowledge expansion, CL allows LLMs to integrate new corpora and incrementally update the knowledge stored in their parameters. This ensures that LLMs remain adaptable, relevant, and effective as facts, domains, languages, and user preferences evolve over time.



In the era of LLMs, the training of language models usually includes multiple stages: pretraining, instruction tuning, preference alignment, and potentially fine-tuning on a downstream task \citep{shi_continual_2024}. Depending on the stage, continual learning can be categorized into continual pretraining, continual instruction tuning, continual preference alignment, and continual end-task learning \citep{ke_continual_2023, shi_continual_2024}. For knowledge expansion, the focus lies on continual pretraining (CPT) and continual preference alignment (CPA). In contrast, continual instruction tuning and continual end-task learning primarily aim to sequentially fine-tune pretrained LLMs for acquiring new skills and solving new tasks, which fall outside the scope of this survey. 
 
In the following sections, we review existing studies that leverage continual pretraining for updating facts, adapting domains, and expanding languages, and continual alignment for updating user preferences.

\subsection{Continual Pretraining for Updating Facts}
\label{sec:continual-pretraining-facts}

This line of research focuses on updating a language model’s outdated internal factual knowledge by incrementally integrating up-to-date world knowledge \citep{jang_towards_2022, ke_continual_2023}. 

Early studies \citep{sun_ernie_2020, rottger-pierrehumbert-2021-temporal-adaptation, lazaridou2021mind, tacl_a_00459} empirically analyze continual pretraining on temporal data, demonstrating its potential for integrating new factual knowledge. \citet{jin-etal-2022-lifelong-pretraining} and \citet{jang_towards_2022} apply traditional continual learning methods to factual knowledge updates in LLMs, evaluating their effectiveness in continual knowledge acquisition. Similarly, \citet{jang_towards_2022} and \citet{kim-etal-2024-carpe} classify world knowledge into time-invariant, outdated, and new categories --- requiring knowledge retention, removal, and acquisition, respectively --- and benchmark existing continual pretraining methods for knowledge updates. 

Additionally, \citet{hu-etal-2023-meta} introduce a meta-trained importance-weighting model to adjust per-token loss dynamically, enabling LLMs to rapidly adapt to new knowledge. \citet{yu-ji-2024-information} investigate self-information updating in LLMs through continual learning, addressing exposure bias by incorporating fact selection into training losses. 

\subsection{Continual Pretraining for Domain Adaptation}
\label{sec:continual-pretraining-domain}

Continual domain adaptative pretraining \citep{ke-etal-2022-continual, ke_continual_2023, wu2024continual} focuses on incrementally adapting an LLM using a sequence of unlabeled, domain-specific corpora. The objective is to enable the LLM to accumulate knowledge across multiple domains while mitigating catastrophic forgetting \citep{mccloskey1989catastrophic} of previously acquired domain knowledge or general language understanding.



\citet{gururangan-etal-2020-dont}  introduced the term of domain-adaptive pretraining, demonstrating that a second phase of pretraining on target domains can effectively update an LLM with new domain knowledge. It is important to note that further pretraining can lead to catastrophic forgetting of general concepts by overwriting essential parameters. To mitigate this, recent works utilize \textit{parameter-isolation} methods which allocate different parameter subsets to distinct tasks or domains and keep the majority of parameters frozen \cite{razdaibiedina2023progressive, wang-etal-2024-learn, wang-etal-2024-rehearsal}. 
DEMix-DAPT \citep{gururangan-etal-2022-demix} replaces every feed-forward network layer in the Transformer model with a domain expert mixture layer, containing one expert per domain. When acquiring new knowledge, only the newly added expert is trained while all others remain fixed. \citet{qin-etal-2022-elle} propose ELLE for efficient lifelong pretraining on various domains. ELLE starts with a randomly initialized LLM and expands the PLM's width and depth to acquire new knowledge more efficiently. \citet{ke-etal-2022-continual} introduce a continual pretraining system which inserts continual learning plugins to the frozen pretrained language models that mitigate catastrophic forgetting while effectively learn new domain knowledge. Similarly, Lifelong-MoE \citep{pmlr-v202-chen23aq} expands expert capacity progressively, freezing previously trained experts and applying output-level regularization to prevent forgetting. 

In a later work, \citet{ke_continual_2023} apply regularization to penalize changes to critical parameters learned from previous data, preventing catastrophic forgetting. Their approach computes the importance of LLM components, such as attention heads and neurons, in preserving general knowledge, applying soft-masking and contrastive loss during continual pretraining to maintain learned knowledge while promoting knowledge transfer.

\subsection{Continual Pretraining for Language Expansion}
\label{sec:continual-pretraining-language}

Continual pretraining (CPT) has emerged as a pivotal strategy for adapting LLMs to new languages, or enhancing performance in underrepresented languages without full retraining \citep{wu2024continual}. Below, we discuss two major areas of expansion enabled by CPT: \textit{natural language expansion} and \textit{programming language expansion}.

\paragraph{Natural Language Expansion.} 

Several recent studies have demonstrated the effectiveness of CPT in expanding language coverage. 
Glot500 \citep{imanigooghari-etal-2023-glot500} and EMMA-500 \citep{ji2024emma} enhance multilingual capabilities using CPT and vocabulary extension. Glot500, based on XLM-R \citep{ruder-etal-2019-unsupervised}, and EMMA-500, built on LLaMA 2 \citep{touvron2023llama}, expand language support up to 500 languages using extensive multilingual corpora.
Similarly, Aya \citep{ustun-etal-2024-aya} applies continual pretraining to the mT5 model \citep{xue-etal-2021-mt5} using a carefully constructed instruction dataset, achieving improved performance across 101 languages. Furthermore, LLaMAX \citep{lu-etal-2024-llamax} enhances multilingual translation by applying continual pretraining to the LLaMA model family. Supporting over 100 languages, it improves translation quality and promotes language inclusivity.

While covering many languages, many multilingual models exhibit suboptimal performance on medium- to low-resource languages \citep{ruder-etal-2019-unsupervised, touvron2023llama, imanigooghari-etal-2023-glot500}.
To bridge this performance gap, researchers have focused on expanding training corpora and strategically applying continual pretraining to enhance the multilingual capabilities of LLMs. \citet{alabi-etal-2022-adapting}, \citet{wang-etal-2023-nlnde}, \citet{fujii2024continual}, and \citet{zhang-etal-2024-aadam} show that continual pretraining on one or more specific languages significantly improves performance across related languages. \citet{blevins-etal-2024-breaking} extend this approach to the MoE paradigm for better parameter efficiency, while \citet{zheng-etal-2024-breaking} investigate scaling laws for continual pretraining by training LLMs of varying sizes under different language distributions and conditions. Additionally, \citet{tran2020english}, \citet{minixhofer-etal-2022-wechsel}, \citet{dobler-de-melo-2023-focus}, \citet{liu-etal-2024-ofa}, and \citet{minixhofer2024zero} explore advanced tokenization and word embedding techniques to further improve LLMs’ multilingual performance in low-resource settings.

\paragraph{Programming Language Expansion.} Going beyond natural languages, continual pretraining has demonstrated significant potential in enhancing the capabilities of LLMs for understanding and generating programming languages.

CERT, proposed by \citet{zancert}, addresses the challenges of library-oriented code generation using unlabeled code corpora. It employs a two-stage framework to enable LLMs to effectively capture patterns in library-based code snippets. CodeTask-CL \citep{yadav-etal-2023-exploring} offers a benchmark for continual code learning, encompassing a diverse set of tasks such as code generation, summarization, translation, and refinement across multiple programming languages. 
Furthermore, continual pretrained models specifically for code understanding and programming from natural language prompts emerged with LLMs, such as Code\-LLaMA \citep{grattafiori2023code}, Llama Pro \citep{wu-etal-2024-llama}, CodeGemma \citep{team2024codegemma} and StarCoder 2 \citep{lozhkov2024starcoder}, consistently outperform general-purpose LLMs of comparable or larger size on code benchmarks.



\subsection{Continual Preference Alignment}
\label{sec:continual-alignment}

Preference alignment ensures that large language models generate responses consistent with human values, improving usability, safety, and ethical behavior. While techniques like Reinforcement Learning from Human Feedback (RLHF) \citep{ziegler2019fine, lambert2022illustrating} align LLMs with static preferences, societal values evolve, requiring continual preference alignment (CPA). It enables LLMs to adapt to emerging preferences while preserving previously learned values, ensuring relevance, inclusivity, and responsiveness to shifting societal expectations. Despite its importance, CPA remains a relatively underexplored area. Below, we briefly discuss two representative works that highlight the potential of this approach: 

\citet{zhang2023copf} propose a non-reinforcement learning approach for continual preference alignment in LLMs. Their method uses function regularization by computing an optimal policy distribution for each task and applying it to regularize future tasks, preventing catastrophic forgetting while adapting to new domains. This provides a single-phase, reinforcement learning-free solution for maintaining alignment across diverse tasks. \citet{zhang2024cppo} introduce Continual Proximal Policy Optimization (CPPO), integrating continual learning into the RLHF framework to accommodate evolving human preferences. CPPO employs a sample-wise weighting strategy to balance policy learning and knowledge retention, consolidating high-reward behaviors while mitigating overfitting and noise. 

As the demand for responsive and inclusive AI grows, CPA is key to keeping LLMs ethical and aligned with evolving user needs, requiring further research to reach its full potential.

\subsection{Applicability and Limitations}
Continual learning is a versatile framework for expanding LLM knowledge across facts, domains, languages, and preferences. It excels in large-scale knowledge integration, retaining previously learned knowledge, 
making it well-suited for tasks like domain adaptation and language expansion \citep{bu2021gaia, jin-etal-2022-lifelong-pretraining, cossu2024continual}. 

However, CL has notable limitations, including a lack of precise control compared to model editing (cf.\ Section~\ref{sec:model-editing}) and retrieval-based methods (cf.\ Section~\ref{sec:retrieval}), inefficiency due to the computational demands of retraining, and limited applicability in black-box models. 
These challenges highlight the 
need for alternative approaches 
like model editing and retrieval, which offer more targeted and efficient updates. 


\section{Model Editing} 
\label{sec:model-editing}

Model editing offers a controllable and efficient solution to update factual knowledge and user preferences in LLMs.
Introduced by \citet{zhu_modifying_2020}, \citet{de-cao-etal-2021-editing} and \citet{mitchell_fast_2022}, it aims at modifying the model’s predictions for specific inputs without affecting unrelated ones.

\citet{yao-etal-2023-editing} and \citet{zhang2024comprehensive} define four key evaluation metrics for model editing: \textbf{(1) reliability}, ensuring the edited model produces the target prediction for the target input; \textbf{(2) generalization}, requiring the edited knowledge to apply to all in-scope inputs — inputs that are directly related to the target input, including rephrasings and semantically similar variations; \textbf{(3) locality}, preserving original outputs for unrelated out-of-scope inputs; and \textbf{(4) portability}, extending the generalization metric by assessing how well updated knowledge transfers to complex rephrasings, reasoning chains, and related facts. 

While recent works \citep{mitchell_memory_2022, madaan-etal-2022-memory, zhong-etal-2023-mquake, zheng-etal-2023-edit} use \textit{model editing} and \textit{knowledge editing} interchangeably for updating factual knowledge, we distinguish between them: model editing is a subset of knowledge editing that modifies model parameters, whereas retrieval-based methods update knowledge dynamically without altering the model’s parameters (see Section~\ref{sec:retrieval}).

\subsection{Model Editing for Updating Facts}
\label{sec:model-editing-facts}

To address outdated or incorrect information \citep{lazaridou2021mind}, model editing research focuses on selectively modifying this knowledge. Below, we highlight key works in this area.

KnowledgeEditor \citep{de-cao-etal-2021-editing} uses a hypernetwork to predict parameter shifts for modifying a fact, trained via constrained optimization for locality. 
Similarly, MEND \citep{mitchell_fast_2022} trains a hypernetwork per LLM layer and decomposes the fine-tuning gradient into a precise one-step parameter update.
Given the findings that feed-forward layers in transformers function as key-value memories \citep{geva-etal-2021-transformer}, \citet{dai-etal-2022-knowledge} introduce a knowledge attribution method to identify these neurons and directly modify their values 
via knowledge surgery.

Recent works employ a locate-and-edit strategy for precise model editing. Using causal tracing, \citet{meng_locating_2022} identify middle-layer feed-forward networks as key to factual predictions and propose ROME, which updates facts by solving a constrained least-squares problem in the MLP weight matrix. MEMIT \citep{meng_mass-editing_2023} extends ROME to modify thousands of facts simultaneously across critical layers while preserving generalization and locality. BIRD \citep{ma2023untying} introduces bidirectional inverse relationship modeling
to mitigate the reverse curse \citep{berglund2023reversal} in model editing. While editing FFN layers has proven effective, PMET \citep{li2024pmet} extends editing to attention heads, achieving improved performance. 
\citet{wang-etal-2024-editing} further shift the focus to conceptual knowledge, using ROME and MEMIT to alter concept definitions,
finding that concept-level edits are reliable but have limited influence on concrete examples.

\subsection{Model Editing for Updating Preferences}
\label{sec:model-editing-preference}

Recent works expand model editing beyond factual corrections to aligning LLMs with user preferences, such as ensuring safety,
reducing bias, 
and preserving privacy
. 

\citet{wang-etal-2024-detoxifying} use model editing to detoxify LLMs, ensuring safe responses to adversarial inputs 
and preserving general LLM capabilities, such as fluency, knowledge question answering, and content summarization. 
Their results show that model editing is promising for detoxification but slightly affects general capabilities. 
Since LLMs can exhibit social biases \citep{gallegos-etal-2024-bias}, \citet{chen_large_2024} propose fine-grained bias mitigation via model editing. Inspired by \citet{meng_locating_2022}, they identify key layers responsible for biased knowledge and insert a feed-forward network to adjust outputs with minimal parameter changes, ensuring generalization, locality, and scalability.
For privacy protection, \citet{wu-etal-2023-depn} extend \citet{dai-etal-2022-knowledge}'s work by identifying privacy neurons that store sensitive information. Using gradient attribution, they deactivate these neurons, reducing private data leakage while preserving model performance.  
Moreover, \citet{mao_editing_2024} apply model editing techniques like MEND to modify personality traits in LLMs, aligning responses to opinion-based questions with target personalities. While effective, this approach can degrade text generation quality.


\subsection{Applicability and Limitations}

Model editing complements continual learning by allowing fine-grained knowledge updates with lower computational costs.
However, research has primarily focused on structured, relational, and instance-level knowledge, with limited exploration of other knowledge types, multilingual generalization, and cross-lingual transfer \citep{nie2024bmike, wei-etal-2025-mlake}.

Additionally, model editing faces several technical challenges, including limited locality and gradual forgetting in large-scale edits \citep{bu2019deep, mitchell_memory_2022, gupta-etal-2024-model, li2024should}, making it more suitable for minor updates.
Additionally, it can impact general LLM capabilities \citep{gu-etal-2024-model, wang-etal-2024-better} and downstream performance \citep{gupta-etal-2024-model}, potentially causing model collapse \citep{yang-etal-2024-fall}. Addressing these issues will enhance model editing’s role alongside continual learning and retrieval, ensuring greater precision in dynamic knowledge adaptation.


\section{Retrieval-based Methods}
\label{sec:retrieval}

Continual learning and model editing modify a model’s parameters to update its internal knowledge, making them implicit knowledge expansion methods \citep{zhang-etal-2023-large}. In contrast, retrieval-based methods \citep{lewis2020retrieval} explicitly integrate external knowledge, allowing models to overwrite outdated or undesired information without parameter modifications. These methods leverage external sources, such as databases, off-the-shelf retriever systems, or the Internet, and thus provide up-to-date or domain-specific knowledge \citep{zhang-etal-2023-large}, making them effective for factual updates and domain adaptation.


\subsection{Retrieval-based Methods for Updating Facts}
\label{sec:retrieval-facts}

Retrieval-based methods enhance LLMs by pairing them with an updatable datastore, ensuring access to current factual information. An early approach, retrieval-augmented generation (RAG) \citep{lewis2020retrieval}, fine-tunes a pre-trained retriever end-to-end with the LLM to improve knowledge retrieval. Similarly, kNN-LM \citep{khandelwal_generalization_2020} interpolates the LLM’s output distribution with k-nearest neighbor search results from the datastore, with later works optimizing efficiency \citep{he-etal-2021-efficient, alon_retomaton_2022} and adapting it for continual learning \citep{peng_semiparametric_2023}.

For factual knowledge editing, \citet{tandon-etal-2022-learning} store user feedback for post-hoc corrections, while \citet{mitchell_memory_2022}, \citet{madaan-etal-2022-memory}, and \citet{dalvi-mishra-etal-2022-towards} retrieve stored edits to guide responses. \citet{chen-etal-2024-robust} introduce relevance filtering to efficiently handle multiple edits. Retrieval-based in-context learning \citep{zheng-etal-2023-edit, ram-etal-2023-context, mallen-etal-2023-trust, yu-etal-2023-augmentation, shi-etal-2024-replug, bi2024decoding} enables dynamic factual updates. 

For complex reasoning, retrieval supports multi-hop question answering and iterative prompting: \citet{zhong-etal-2023-mquake} propose iterative prompting for multi-hop knowledge editing, while \citet{gu-etal-2024-pokemqa} use a scope detector to retrieve relevant edits and improve question decomposition via entity extraction and knowledge prompts. Similarly, \citet{shi_retrieval_2024} enhance multi-hop question answering by retrieving fact chains from a knowledge graph with mutual information maximization and redundant fact pruning. 

In multi-step decision-making, retrieval is combined with Chain-of-Thought (CoT) reasoning \citep{trivedi-etal-2023-interleaving, press-etal-2023-measuring}. Retrieval also aids post-generation fact-checking and refinement \citep{gao-etal-2023-rarr, peng2023checkfactstryagain, song_knowledge_2024} by revising generated text or prompts based on retrieved facts.



For a more comprehensive review of retrieval-based factual knowledge updates, we refer to \citet{zhang-etal-2023-large}.



\subsection{Retrieval-based Methods for Domain Adaptation}
\label{sec:retrieval-domain}
Retrieval-based methods have been widely adopted for various domain-specific tasks, e.g., in science and finance. By integrating retrieved external knowledge, these models enhance their adaptability to specialized domains, improving decision-making, analysis, and information synthesis. 

In the biomedical domain, retrieval-based approaches facilitate tasks, such as molecular property identification and drug discovery by integrating structured molecular data and information about biomedical entities like proteins, molecules, and diseases \citep{wang2023retrieval, liu2023multi, wang2024biobridge, yang2024promptbased}. For instance, \citet{wang2023retrieval} and \citet{li2024empowering} introduce retrieval-based frameworks that extract relevant molecular data from databases to guide molecule generation. In protein research, retrieval-based approaches enhance protein representation and generation tasks \citep{ma-etal-2024-retrieved, sun2023graphvf}. Additionally, \citet{lozano2023clinfo} develop a clinical question-answering system that retrieves relevant biomedical literature to provide more accurate responses in medical contexts. 

The finance domain, characterized by its data-driven nature, also benefits from retrieval-based methods \citep{li2024incorporating, li2024knowledge}. \citet{zhang2023enhancing} enhance financial sentiment analysis by retrieving real-time financial data from external sources. 
Furthermore, financial question-answering also benefits from retrieval-based methods, which involves extracting knowledge from professional financial documents. \citet{lin2024revolutionizing} introduces a PDF parsing method integrated with retrieval-augmented LLMs to retrieve relevant financial insights. 

\subsection{Applicability and Limitations}
\label{sec:retrieval-discussion}
Despite their advantages, retrieval-based methods also come with several limitations. A major challenge is their reliance on external knowledge sources, which can introduce inconsistencies or outdated information if not properly curated \citep{jin-etal-2024-bider, xu-etal-2024-knowledge-conflicts}. Their effectiveness also depends on the quality and scope of the retrieval system \citep{bai-etal-2024-mt, liu2024graphsnapshot}; poor indexing or noisy retrieval may lead to irrelevant or misleading information. Another key issue is maintaining knowledge consistency across queries. Since retrieval-based methods do not update model parameters, contradictions can arise between retrieved facts and previously generated responses, affecting coherence 
\citep{njeh2024enhancing, zhao2024dense, li-etal-2024-graphreader}.  

Addressing these challenges is essential to improving retrieval-based approaches and ensuring their seamless integration with other LLM adaptation techniques.

\section{Challenges, Opportunities, Guidelines}

\paragraph{Solving Knowledge Conflicts.}
An inherent challenge of expanding a model’s knowledge is the emergence of knowledge conflicts, which can undermine the consistency and trustworthiness of LLMs \citep{xu-etal-2024-knowledge-conflicts}. Studies have identified various types of conflicts following knowledge updates, including (i) temporal misalignment \citep{luu-etal-2022-time}, where outdated and newly learned facts coexist inconsistently, (ii) model inconsistencies \citep{huang2021factual}, where responses to similar queries vary unpredictably, and (iii) hallucinations \citep{ji2023survey}, where the model generates fabricated or contradictory information. While some efforts have been made to address these issues \citep{zhang-choi-2023-mitigating, mallen-etal-2023-trust, zhou-etal-2023-context, xie2024adaptive}, they remain an open challenge that requires further research and more robust solutions.

\paragraph{Minimizing Side Effects.}
Continual learning and model editing, both of which involve modifying model parameters, inevitably introduce side effects. A major challenge in continual learning is catastrophic forgetting \citep{mccloskey1989catastrophic}, where newly acquired knowledge overwrites previously learned information. In LLMs, the multi-stage nature of training exacerbates this issue, leading to cross-stage forgetting \citep{wu2024continual}, where knowledge acquired in earlier stages is lost as new training phases are introduced.
For model editing, recent studies have shown that large-scale edits, particularly mass edits, can significantly degrade the model’s general capabilities, such as its language modeling performance \citep{wang-etal-2024-better} or accuracy on general NLP tasks \citep{li2024unveiling, li2024should, wang2024missing}.
Effectively addressing these challenges is crucial for maximizing the potential of these methods for large-scale knowledge expansion while maintaining model stability and overall performance.

\paragraph{Comprehensive Benchmarks.}
Although this paper explores the properties, strengths, and weaknesses of various methods for knowledge expansion, the discussion remains largely theoretical due to the lack of a comprehensive benchmark datasets for a uniform evaluation and a proper comparison. Existing works, such as \citet{jang_towards_2022}, \citet{liska2022streamingqa}, and \citet{kim-etal-2024-carpe}, provide factual knowledge-based datasets and evaluate continual pretraining and/or retrieval-based methods. However, their experiments are limited in scale and fail to offer a comprehensive assessment. Developing benchmarks that encompass a variety of knowledge types and enable the evaluation of all methods would provide a more holistic and systematic understanding of their relative effectiveness.

\paragraph{General Guideline.}
Selecting the appropriate method for knowledge expansion in LLM depends on the application context and type of knowledge that needs to be updated. 

(\romannumeral1) For \textbf{factual knowledge}, model editing is ideal for precise, targeted updates, such as correcting specific facts, due to its efficiency and high level of control. Retrieval-based methods are effective for integrating dynamic or frequently changing facts, as they allow updates without modifying the model’s parameters, making them suitable for black-box applications. For large-scale factual updates, continual learning is preferred as it enables the incremental integration of new knowledge while preserving previously learned information.

(\romannumeral2) For \textbf{domain knowledge}, both continual learning and retrieval-based methods are applicable. Continual learning excels in large-scale adaptation, using domain-specific corpora to ensure the model retains general knowledge while adapting to specialized contexts. Retrieval-based methods complement this by dynamically providing domain-specific information without requiring model modifications, making them valuable in scenarios where static updates are impractical.

(\romannumeral3) For \textbf{language knowledge}, continual learning is the only method capable of supporting large-scale language expansion. It facilitates the integration of multilingual corpora and provides the foundational updates necessary for underrepresented or low-resource languages. 

(\romannumeral4) For \textbf{preference updates}, such as aligning models with evolving user values or ethical norms, continual alignment is typically achieved by combining continual learning techniques with preference optimization methods, such as reinforcement learning from human feedback. These approaches enable models to dynamically adapt to changing preferences while retaining alignment with previously learned values.

\textbf{Summary.} \textbf{Continual learning} is indispensable for large-scale updates like domain adaptation and language expansion, where foundational and incremental updates are required. \textbf{Model editing} excels at precise factual updates, while \textbf{retrieval-based methods} offer dynamic access to factual and domain knowledge without altering the model. A well-informed selection or combination of these methods ensures efficient and effective knowledge expansion tailored to specific use cases.

\section{Conclusions}
Adapting large language models to evolving knowledge is essential for maintaining their relevance and effectiveness. This survey explores three key adaptation methods --- continual learning for large-scale updates, model editing for precise modifications, and retrieval-based approaches for external knowledge access without altering model parameters. We examine how these methods support updates across factual, domain-specific, language, and user preference knowledge while addressing challenges like scalability, controllability, and efficiency. By consolidating research and presenting a structured taxonomy, this survey provides insights into current strategies and future directions, promoting the development of more adaptable and efficient large language models.

\section*{Limitations}
This survey provides a comprehensive overview of knowledge expansion techniques for LLMs. However, due to page constraints, we had to limit its scope and prioritize certain aspects: 

First, the paper only provides a high-level overview of each method rather than an in-depth analysis. 
This can limit the understanding of the nuances and specific applications of each technique, as well as implementation details. 

Second, our work is a literature review of adaptation methods rather than an empirical study evaluating their actual performance. While we analyze existing strategies, we do not benchmark or experimentally compare their effectiveness, leaving room for future studies to assess their practical impact under real-world conditions.

Third, we focus solely on text-based models and do not cover vision-language models, which integrate multi-modal learning for textual and visual understanding. While the methods covered in this survey could be used to adapt the language encoders of such models in theory, extending these adaptation methods to vision-language models remains an open research direction. 

Finally, this survey reflecting the current state of research might become outdated as new research is published, as the field of LLMs is rapidly evolving and new methods for knowledge expansion are continuously being developed.


\bibliography{anthology,custom}

\newpage
\clearpage
\appendix

\section{Appendix}
\label{sec:appendix}

\subsection{Comprehensive Taxonomy of Methods}
\label{sec:appendix_taxonomy}
\FloatBarrier
\begin{figure}[H]
\centering
\resizebox{\textwidth}{!}{
    \begin{forest}
    for tree={
        text width=2.0cm,
        grow=east,
        parent anchor=east,
        child anchor=west,
        anchor=center,
        rounded corners,
        font=\footnotesize,
        edge={thick},
        forked edges,
        draw,
        l sep=5mm,
        s sep=5mm
    },
    [Knowledge Expansion Methods,
        rotate=90,
        text width=4.2cm,
        draw=gray!50, 
        line width=0.5mm, 
        fill=gray!20,
        text centered,
        [
            Retrieval-based \\Methods \\(§\ref{sec:retrieval}), 
            draw=yellow!50, 
            line width=0.5mm, 
            fill=yellow!20,
            text centered,
            [
                Domain \\Adaptation \\(§\ref{sec:retrieval-domain}), 
                draw=yellow!50, 
                line width=0.5mm, 
                fill=yellow!20,
                text centered,
                [
                    {
                        \citet{khandelwal_generalization_2020},
                        \citet{lewis2020retrieval},
                        \citet{he-etal-2021-efficient},
                        \citet{alon_retomaton_2022}, 
                        \citet{shi-etal-2022-nearest},
                        \citet{wang2023retrieval},
                        \citet{liu2023multi},
                        \citet{lozano2023clinfo},
                        \citet{peng_semiparametric_2023},
                        \citet{ram-etal-2023-context},
                        \citet{sun2023graphvf},
                        \citet{zhang2023enhancing},
                        \citet{li2024empowering},
                        \citet{li2024incorporating},
                        \citet{li2024knowledge},
                        \citet{lin2024revolutionizing},
                        \citet{ma-etal-2024-retrieved},
                        \citet{wang2024biobridge},
                        \citet{yang2024promptbased},
                        \citet{yepes2024financial}
                    },
                    draw=yellow!50,
                    fill=yellow!20,
                    text width=12.0cm
                ]
            ]
            [
                Updating \\Facts \\(§\ref{sec:retrieval-facts}), 
                draw=yellow!50, 
                line width=0.5mm, 
                fill=yellow!20,
                text centered,
                [
                    {
                        \citet{khandelwal_generalization_2020},
                        \citet{lewis2020retrieval},
                        \citet{he-etal-2021-efficient},
                        \citet{alon_retomaton_2022},
                        \citet{dalvi-mishra-etal-2022-towards},
                        \citet{madaan-etal-2022-memory},
                        \citet{mitchell_memory_2022},
                        \citet{tandon-etal-2022-learning},
                        \citet{gao-etal-2023-rarr},
                        \citet{mallen-etal-2023-trust},
                        \citet{peng_semiparametric_2023},
                        \citet{peng2023checkfactstryagain},
                        \citet{press-etal-2023-measuring},
                        \citet{ram-etal-2023-context},
                        \citet{si_prompting_2023},
                        \citet{trivedi-etal-2023-interleaving},
                        \citet{yu-etal-2023-augmentation},
                        \citet{zheng-etal-2023-edit},
                        \citet{zhong-etal-2023-mquake},
                        \citet{bi2024decoding},
                        \citet{chen-etal-2024-robust},
                        \citet{gu-etal-2024-pokemqa},
                        \citet{shi-etal-2024-replug},
                        \citet{shi_retrieval_2024},
                        \citet{song_knowledge_2024}
                    },
                    draw=yellow!50,
                    fill=yellow!20,
                    text width=12.0cm
                ]
            ]
        ]
        [
            Model \\Editing \\(§\ref{sec:model-editing}), 
            draw=red!50, 
            line width=0.5mm, 
            fill=red!20,
            text centered,
            [
                Updating \\Preferences \\(§\ref{sec:model-editing-preference}), 
                draw=red!50, 
                line width=0.5mm, 
                fill=red!20,
                text centered,
                [
                    {
                        \citet{wu-etal-2023-depn}, 
                        \citet{chen_large_2024}, 
                        \citet{mao_editing_2024}, 
                        \citet{patil_2024_can},
                        \citet{wang-etal-2024-detoxifying}
                    },
                    draw=red!50,
                    fill=red!20,
                    text width=12.0cm
                ]
            ]
            [
                Updating \\Facts \\(§\ref{sec:model-editing-facts}), 
                draw=red!50, 
                line width=0.5mm,
                fill=red!20,
                text centered,
                [
                    {
                        \citet{zhu_modifying_2020},
                        \citet{de-cao-etal-2021-editing},
                        \citet{dai-etal-2022-knowledge},
                        \citet{meng_locating_2022},
                        \citet{mitchell_fast_2022},
                        \citet{huang_transformer-patcher_2023},
                        \citet{ma2023untying},
                        \citet{meng_mass-editing_2023},
                        \citet{li2024pmet},
                        \citet{wang-etal-2024-editing}
                    },
                    draw=red!50,
                    fill=red!20,
                    text width=12.0cm
                ]
            ] 
        ]
        [
            Continual \\Learning \\(§\ref{sec:continual-learning}), 
            draw=orange!50, 
            line width=0.5mm, 
            fill=orange!20,
            text centered,
            [
                Continual \\Preference \\Alignment \\(§\ref{sec:continual-alignment}), 
                draw=orange!50, 
                line width=0.5mm, 
                fill=orange!20,
                text centered,
                [
                    {
                        \citet{zhang2023copf},
                        \citet{zhang2024cppo}
                    },
                    draw=orange!50, 
                    fill=orange!20,
                    text width=12.0cm
                ]
            ]
            [
                Continual \\Pretraining, 
                draw=orange!50, 
                line width=0.5mm, 
                fill=orange!20, 
                text centered,
                [
                    Language \\Expansion \\(§\ref{sec:continual-pretraining-language}),
                    draw=orange!50, 
                    line width=0.5mm, 
                    fill=orange!20,
                    text centered,
                    [
                        Programming Language,
                        draw=orange!50, 
                        line width=0.5mm, 
                        fill=orange!20,
                        text centered,
                        [
                            {
                                \citet{zancert},
                                \citet{grattafiori2023code},
                                \citet{yadav-etal-2023-exploring},
                                \citet{lozhkov2024starcoder},
                                \citet{team2024codegemma},
                                \citet{wu-etal-2024-llama}
                            },
                            draw=orange!50, 
                            fill=orange!20,
                            text width=6.4cm
                        ]
                    ]
                    [
                        Natural \\Language,
                        draw=orange!50, 
                        line width=0.5mm, 
                        fill=orange!20,
                        text centered,
                        [
                            {
                                \citet{tran2020english},
                                \citet{minixhofer-etal-2022-wechsel},
                                \citet{alabi-etal-2022-adapting},
                                \citet{wang-etal-2023-nlnde},
                                \citet{imanigooghari-etal-2023-glot500},
                                \citet{dobler-de-melo-2023-focus},
                                \citet{blevins-etal-2024-breaking},
                                \citet{fujii2024continual},
                                \citet{liu-etal-2024-ofa},
                                \citet{lu-etal-2024-llamax},
                                \citet{ji2024emma},
                                \citet{minixhofer2024zero},
                                \citet{ustun-etal-2024-aya},
                                \citet{zhang-etal-2024-aadam},
                                \citet{zheng-etal-2024-breaking},
                            },
                            draw=orange!50, 
                            fill=orange!20,
                            text width=6.4cm
                        ]
                    ]
                ]
                [
                    Domain \\Adaptation \\(§\ref{sec:continual-pretraining-domain}),
                    draw=orange!50, 
                    line width=0.5mm, 
                    fill=orange!20,
                    text centered,
                    [
                        {
                            \citet{gururangan-etal-2020-dont},
                            \citet{gururangan-etal-2022-demix},
                            \citet{ke-etal-2022-continual},
                            \citet{qin-etal-2022-elle},
                            \citet{pmlr-v202-chen23aq},
                            \citet{ke_continual_2023},
                            \citet{razdaibiedina2023progressive},
                            \citet{wang-etal-2024-rehearsal},
                            \citet{wang-etal-2024-learn}
                        },
                        draw=orange!50, 
                        fill=orange!20,
                        text width=9.2cm
                    ]
                ]
                [
                    Updating \\Facts \\(§\ref{sec:continual-pretraining-facts}),
                    draw=orange!50, 
                    line width=0.5mm, 
                    fill=orange!20, 
                    text centered,
                    [
                        {
                            \citet{sun_ernie_2020}, 
                            \citet{jang_towards_2022},
                            \citet{lazaridou2021mind},
                            \citet{rottger-pierrehumbert-2021-temporal-adaptation},
                            \citet{tacl_a_00459},
                            \citet{jin-etal-2022-lifelong-pretraining},
                            \citet{hu-etal-2023-meta},
                            \citet{kim-etal-2024-carpe},
                            \citet{yu-ji-2024-information}
                        },
                        draw=orange!50, 
                        fill=orange!20,
                        text width=9.2cm,
                    ]
                ],
            ] 
        ]
    ]
    \end{forest}
}
\end{figure}

\begin{minipage}{\textwidth}
    
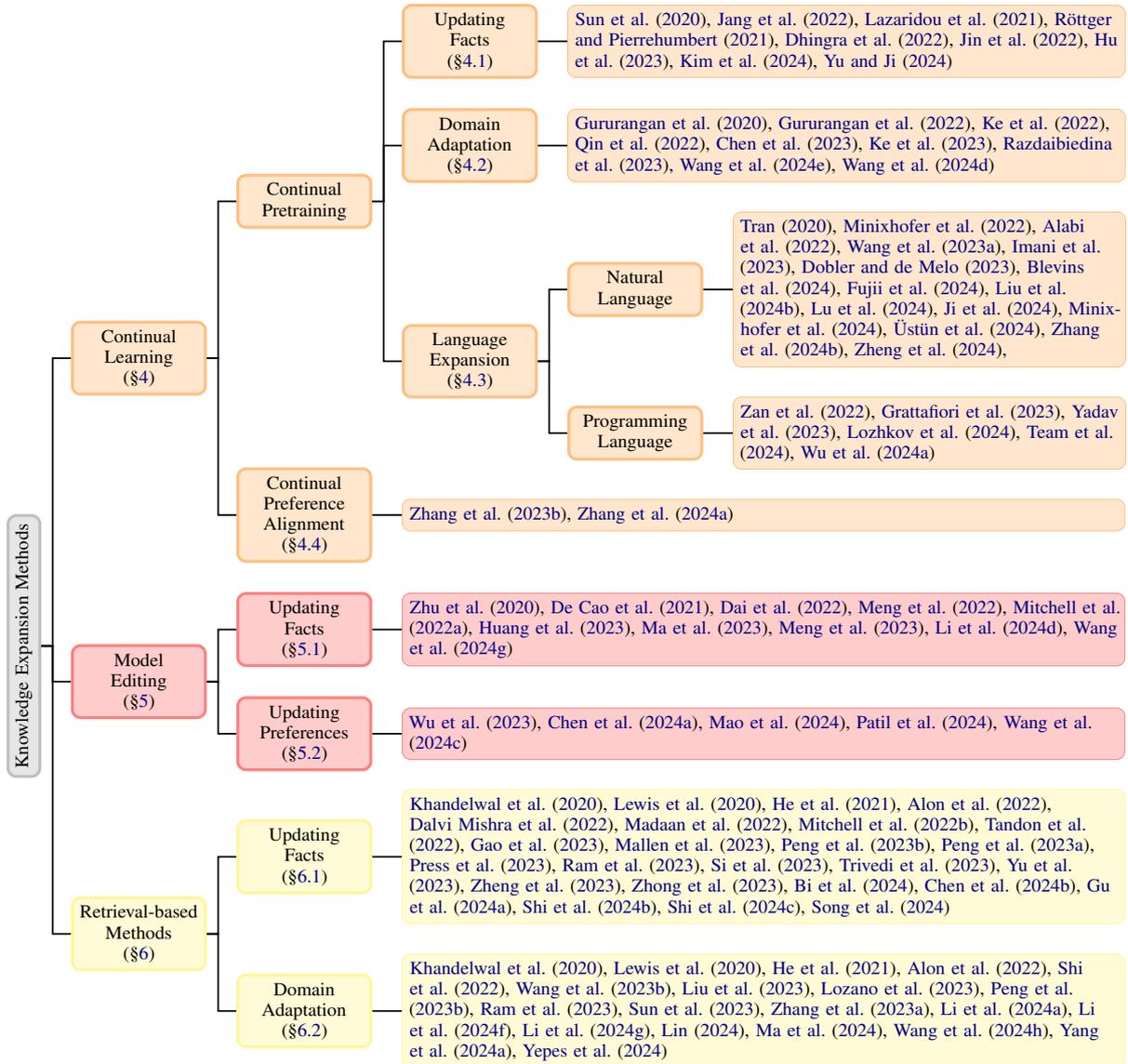
\captionof{figure}{Taxonomy of methods for expanding LLM knowledge.}
    \label{fig:taxonomy_comprehensive}
\end{minipage}

\end{document}